\pdfoutput=1
\documentclass[11pt]{article}

\usepackage{ACL2023}

\usepackage{times}
\usepackage{latexsym}

\usepackage[T1]{fontenc}
\usepackage[utf8]{inputenc}

\usepackage{microtype}

\usepackage{inconsolata}

\usepackage{amsmath}
\usepackage{amsfonts}
\usepackage{amssymb}
\usepackage{graphicx}
\usepackage{multirow}
\usepackage{booktabs}
\usepackage{graphicx}
\usepackage{subcaption}

\newcommand{\sr}[3]{$#1\!#2\!#3$}   % space reduced command

\DeclareMathOperator{\Softmax}{softmax}

\newcommand{\vhead}{\mathcal{V}_{head}(x_{<t})}
\newcommand{\pexp}{p_\textsc{exp} (x_t{\mid}x_{<t})}
\newcommand{\pama}{p_\textsc{ama} (x_t{\mid}x_{<t})}
\newcommand{\acd}{\textit{ACD}}
\newcommand{\med}{\textit{GPT2-Medium}}
\newcommand{\neo}{\textit{GPT-Neo-125M}}
\newcommand{\xl}{\textit{GPT2-XL}}
\newcommand{\wikitext}{WikiText-$103$}

\title{The Benefits of Bad Advice: Autocontrastive Decoding across Model Layers}

\author{\bf{Ariel Gera, Roni Friedman, Ofir Arviv, Chulaka Gunasekara, }\\
\bf{Benjamin Sznajder, Noam Slonim, Eyal Shnarch} \\
\\
IBM Research \\
\{ariel.gera1, ofir.arviv, chulaka.gunasekara\}@ibm.com,
\\\{roni.friedman-melamed, benjams, noams, eyals\}@il.ibm.com}
\begin{document}
\maketitle

\begin{abstract}
Applying language models to natural language processing tasks typically relies on the representations in the final model layer, as intermediate hidden layer representations are presumed to be less informative. In this work, we argue that due to the gradual improvement across model layers, additional information can be gleaned from the contrast between higher and lower layers during inference. Specifically, in choosing between the probable next token predictions of a generative model, the predictions of lower layers can be used to highlight which candidates are best avoided. We propose a novel approach that utilizes the contrast between layers to improve text generation outputs, and show that it mitigates degenerative behaviors of the model in open-ended generation, significantly improving the quality of generated texts. 
% By examining the behavior of different model layers, we also highlight some shifts in language modeling and generation characteristics within a single model. 
Furthermore, our results indicate that contrasting between model layers at inference time can yield substantial benefits to certain aspects of general language model capabilities, more effectively extracting knowledge during inference from a given set of model parameters.

\end{abstract}
\section{Introduction} \label{sec:intro}

For a wide range of natural language processing tasks, the standard practice is to rely on deep neural networks with a transformer-based architecture \citep{vaswani2017attention}. Such models are composed of multiple transformer layers, where typically the representations of the final layer are used for the downstream task.
As shown in prior works, some of the representational knowledge required for performing downstream tasks can already be found within intermediate layers of the model \citep{geva-etal-2021-transformer, geva2022transformer}; at the same time, relying on the representations of lower model layers does result in decreased performance, specifically for inputs that are more challenging \citep{schwartz-etal-2020-right, xin-etal-2020-deebert, elbayad2020depth, sun-etal-2022-simple, schusterconfident, din2023jump}.

% \usepackage{graphicx}
% \begin{table}[]
% \resizebox{\columnwidth}{!}{%
% \begin{tabular}{ll}
% \hline
% \multicolumn{1}{|l|}{\textit{Prompt}}    & \multicolumn{1}{l|}{ֿֿֿֿ\textit{The venues are as follows:}}                                                                                     \\ \hline
% \multicolumn{1}{|l|}{\textbf{Mid Layer}} & \multicolumn{1}{l|}{\begin{tabular}[c]{@{}l@{}}The main entrance is a small town, \\ with a small town\end{tabular}}                \\ \hline
% \multicolumn{1}{|l|}{\textbf{Top Layer}} & \multicolumn{1}{l|}{\begin{tabular}[c]{@{}l@{}}The Royal Albert Hall, London \\ The Royal Albert Hall, London\end{tabular}}         \\ \hline
% \multicolumn{1}{|l|}{\textbf{ACD}}       & \multicolumn{1}{l|}{\begin{tabular}[c]{@{}l@{}}In the heart of London's financial\\ district is The Barbican Theatre.\end{tabular}} \\ \hline
%                                          &                                      

\begin{figure}[t]
\includegraphics[scale=0.25]{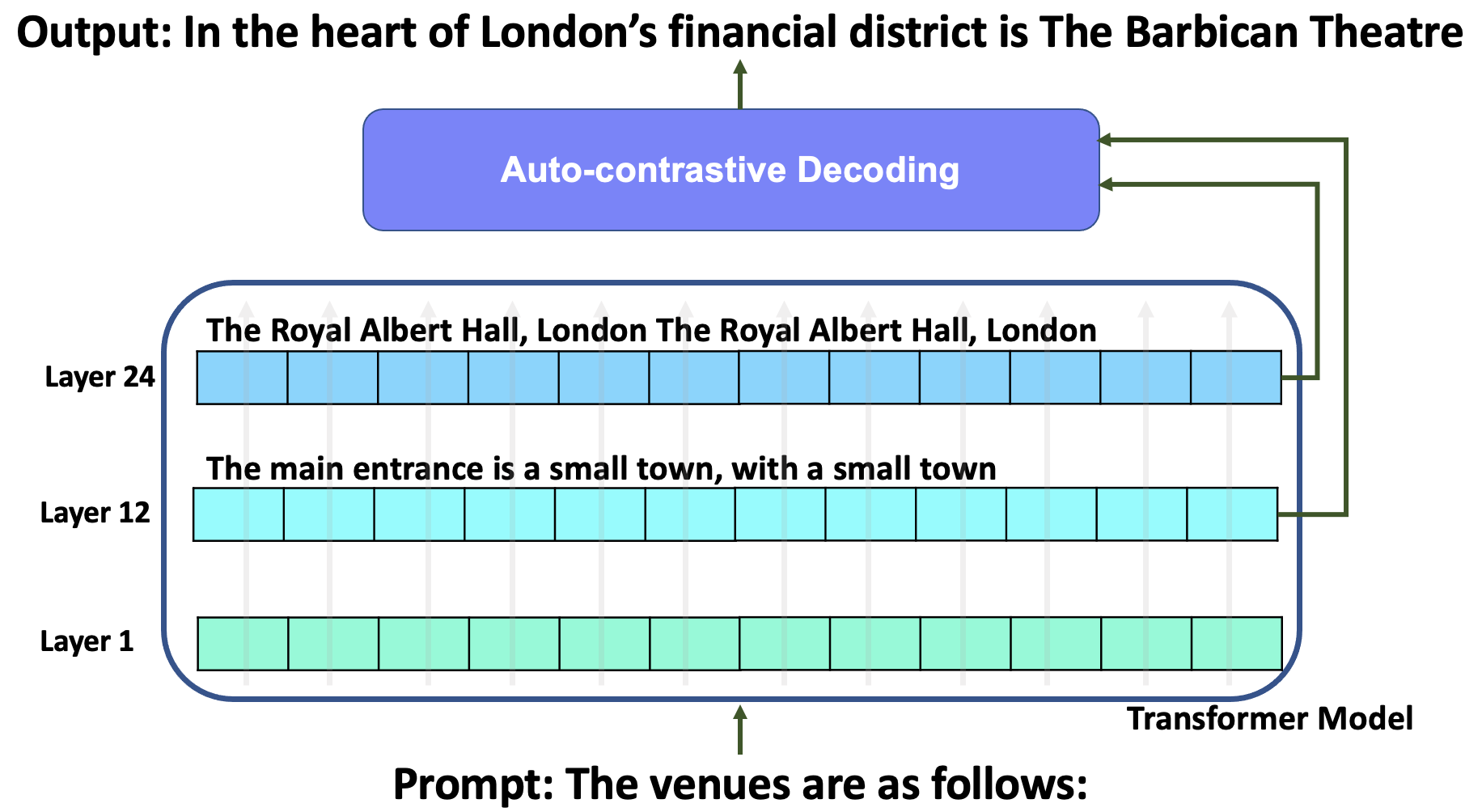}
  \caption{\textbf{An example of auto-contrastive decoding (\acd) with GPT2}, where the top layer ($24$) is taken as the expert and contrasted with layer $12$, the amateur. As decoding is done token by token, we can only see the direct effect on the first token, where \acd{} leads to selecting an alternative high probability token - "In".}
  \label{fig:intro}
\end{figure}

Recently, 
\citet{li2022contrastive} 
considered a scenario involving two language models; one is a very large pre-trained model, termed the \textit{expert}, and the other is a much smaller version of the same architecture, termed the \textit{amateur}.
Importantly, whereas these models share some failure modes and undesirable behaviors, the expert model clearly outperforms the amateur model in language model tasks. Focusing on an open-ended auto-regressive text generation task, they show that it is possible to exploit the contrast between the predictions of the expert and amateur to obtain an improved generated output. They term this method \textit{Contrastive Decoding}. Specifically, they demonstrate that it is sometimes beneficial to prefer predictions to which only the expert model assigns a high probability, versus predictions to which both the expert and the amateur assign high probabilities. Intuitively, since the amateur model has a stronger propensity than the expert for problematic behaviors (e.g., repetitiveness in the case of text generation), we may be able to diminish such behaviors by demoting predictions that are strongly supported by the amateur model.

This scenario relies on a delicate balance: on the one hand, when making a prediction in a relatively simpler context, one would expect both the expert and amateur models to be highly confident about the prediction, and justifiably so; in contrast, where both of them assign very low likelihoods to a certain prediction, these prediction probabilities may be uninformative. Thus, the aim of considering the amateur's predictions during generation is to better inform a choice between a set of relatively plausible predictions given an input; in other words, the predictions of the amateur can serve as a tie-breaker of sorts, helping to highlight which out of a set of plausible alternative predictions is more "expert-like" and less "amateur-like".

Inspired by \citet{li2022contrastive}, in this work we ask whether within a \textit{single} language model, intermediate hidden layers can similarly be viewed as ``amateur'' versions of the final ``expert'' output layer. Given indications that model representations gradually improve as an input progresses through its layers \citep{elbayad2020depth, geva2022transformer}, we aim to examine whether the contrast or gap between the outputs at different model layers can be harnessed to obtain better generation predictions.
In other words, we posit that the sub-optimal predictions of intermediate hidden layers carry additional information, which can be utilized during inference to obtain more desirable next-token predictions.

Our approach, which we term \textit{Auto-contrastive Decoding (ACD)}, redistributes a given model's probability distribution for the next token, by maximizing the difference between the log-probabilities of the final layer and those of an intermediate hidden layer. This setting, where the expert and amateur are situated within the same language model, and their predictions can be carefully contrasted at inference, is a highly practical one and can be easily applied to language models of different sizes.

Our results show that \acd{} enables getting significantly better predictions out of a given language model, without changing its pre-trained weights.

Figure \ref{fig:intro} illustrates an example of \acd{} applied to GPT2, considering layer $12$ as the amateur and layer $24$ as the expert. Both layers exhibit repetitiveness, but applying \acd{} generates a much improved output altogether.

The main contributions of this work are as follows:

1. We reproduce the findings of \citet{li2022contrastive} using a \textit{single medium-size} model, by suggesting a novel intra-model auto-contrastive setting.  

2. We demonstrate that \acd{} improves some aspects of language generation capabilities of pre-trained language models, in essence extracting more knowledge from the model at inference time. We present human evaluation results showing that this brings it to par with larger language models.

3. We release our code and the pre-trained model checkpoints used for experiments in this paper, in order to facilitate further research in this area\footnote{\url{https://github.com/IBM/auto-contrastive-generation}}.

\section{Related Work}
There have been a number of studies on analyzing the characteristics of different layers of transformer models. \citet{rogers2020primer, van2019does} used probing to report that in BERT models the lower layers carry the most information about linear word order, syntactic information is most prominent in the middle layers, and the final layers of BERT are the most task-specific. \citet{van2019does} also show that similar behavior is observed in other transformer models such as GPT2. \citet{geva-etal-2021-transformer, geva2022transformer} studied the role of feed-forward layers in transformer models. They demonstrate that representations across different layers capture meaningful semantic and syntactic patterns, and describe how model predictions are gradually refined as they progress across the different layers.

Aiming to reduce the computational load of transformers, multiple works have explored early-exiting, i.e., performing some calculations without passing through all of the model layers. Such works allow for an early (fast) `exit' from neural network calculations -- for simple instances that can be solved with high accuracy by lower layers -- while using a late (slow) `exit' for more challenging instances \cite{simoulin-crabbe-2021-many,schwartz-etal-2020-right, xin-etal-2020-deebert, elbayad2020depth, sun-etal-2022-simple, schusterconfident}. 

Decoding algorithms are commonly classified as search-based and sampling-based. Search-based methods \cite{steinbiss1994improvements} optimize for the language model log-probabilities, while sampling methods \cite{holtzman2019curious,fan-etal-2018-hierarchical} draw the next token from a truncated distribution. The idea of using contrast during decoding has been explored in several studies. \citet{liu2021dexperts} combine a pretrained LM with `expert' LMs and `anti-expert' LMs, where tokens only get high probability if they are considered likely by the experts and unlikely by the anti-experts. \citet{su2022contrastive} propose constrastive search for decoding, where the generated output is selected from the set of most probable candidates predicted by the model while being discriminative with respect to the context. More recently, \citet{li2022contrastive} suggested to contrast between the likelihood under a large LM (expert) and a small LM (amateur) during decoding. The present work differs significantly from the aforementioned contrastive approaches, in that we contrast the next-token distributions within a single LM, across expert and amateur layers.

\section{Auto-contrastive Decoding}
We set the goal of applying the \textbf{Contrastive Decoding} (CD) method, from \citet{li2022contrastive}, using a \textit{single} model rather than two different models (in their setting, a large and a small model of the same model architecture). As mentioned in the introduction, this setting is more practical and less computationally demanding.
Thus, we generally follow the CD approach to calculate the next-token predictions, by contrasting the predictions of the expert with those of the amateur. However, in our setting, both the expert and the amateur are situated in the same model, and are defined by two different \textit{layers} of that model. We term this new method \textbf{Auto-contrastive Decoding} (ACD). 

Next, we describe how we obtain the expert and the amateur from a single model;  and in \S\ref{ssec:autocontrastive}, we define the auto-contrastive next-token distribution, given the probability distributions of the expert and the amateur.

\subsection{Expert and Amateur in One Model} \label{ssec:heads}
Given a pre-trained language model, \textit{LM$_{\textrm{orig}}$}, we take its final output layer as the \textit{expert}. Similar to \citet{li2022contrastive}, we denote $\pexp$ as the next-token probability distribution of this layer, conditioned on the preceding context ($x_t$ being the next token to predict, and $x_{<t}$ is the context that precedes it).

To obtain the \textit{amateur} from the same model, we add a linear head to one of its intermediate hidden layers, making LM$_{\textrm{orig}}$ a multi-exit model \citep{scardapane2020should, liu-etal-2022-towards-efficient}. This new head maps the output of the intermediate layer, given a preceding context, to a probability distribution over the vocabulary for the next token, denoted $\pama$.

To train only this new head, we freeze all of the existing pre-trained weights of LM$_{\textrm{orig}}$; we then train the model, applying the same self-supervised objective that was used to pre-train LM$_{\textrm{orig}}$.

In this training we do not aim to fully reproduce the original pre-training of LM$_{\textrm{orig}}$; note that we are training a relatively small number of parameters, and thus can use less data and perform fewer training steps. This reduced training is likely to lead to certain disparities between the amateur head and the expert head, as the latter was trained as part of the original LM$_{\textrm{orig}}$ pre-training. 
Thus, we also train a new expert head, using an identical procedure as the one used to train the amateur head\footnote{Our motivation for training a new expert head was to explore a scientifically "cleaner" scenario, where there is a more straightforward relation between the amateur and expert head. However, based on the results we report in App.~\ref{app:original_head}, from a practical standpoint this may not be necessary.}.

To amplify the performance gap between the expert and the amateur, \citet{li2022contrastive} introduced another limitation on the amateur model (apart from it being a small version of the expert model) -- the preceding context given to the amateur model is restricted, notably shorter than the one provided to the expert model. In \acd{} we opt to abstain from this additional (and somewhat arbitrary) limitation, and both $\pexp$ and $\pama$ are conditioned on the same full context. 

\subsection{Auto-contrastive Next-token Distribution} 
\label{ssec:autocontrastive}
Next, we describe the auto-contrastive decoding, ACD. This method outputs a token-level probability distribution by contrasting the next-token distribution of the expert, $\pexp$, with that of the amateur, $\pama$. 

Following \citet{li2022contrastive}, we first implement the CD adaptive plausibility constraint, $\vhead$, defined by: 
\begin{align}
\label{eq:constraint}
& \vhead = \\ \nonumber
& \{x_t\!\in\!\mathcal{V}: \pexp \geq \alpha \max_{x'_t\in \mathcal{V}} p_\textsc{exp} (x'_t{\mid}x_{<t}) \} 
\end{align}
Given a preceding context $x_{<t}$, this constraint selects a subset of plausible next tokens, out of the vocabulary $\mathcal{V}$, whose probabilities are above a threshold. The threshold is a fraction $\alpha$ of the probability of the token with the highest probability in the vocabulary. The hyperparameter $\alpha$ is in the range $[0, 1]$, and it is set to $0.1$ in all our experiments, as done by \citet{li2022contrastive}.

The score for a plausible $x_t$, i.e., \sr{x_t}{\in}{\vhead},  indicating its likelihood to be the next token given the context $x_{<t}$, is calculated by contrasting the probabilities given to it by the expert and by the amateur:
\begin{equation}
S(x_t{\mid}x_{<t}) = \log \pexp - \log \pama
\label{eq:cd}
\end{equation}

Note that this contrastive score is only applied to the tokens in $\vhead$. 
This constraint serves an important purpose in that it helps avoid assigning high probabilities to very unlikely tokens, 
namely those for which $p_\textsc{exp}$ is very low; at the same time, where the expert is highly confident about a single top prediction, it helps ensure that $p_\textsc{ama}$ does not alter the final outcome\footnote{Consider for example a case where the token with the maximum probability is assigned a very high probability, e.g., $\max_{x'_t\in \mathcal{V}} p_\textsc{exp} (x'_t{\mid}x_{<t})>0.9$, and where $p_\textsc{ama}$ for this token is also quite high. In this scenario, while Eq.~\ref{eq:cd} may give a very low contrast score $S(x'_t{\mid}x_{<t})$, this will be the only token that meets $\vhead$ (Eq.~\ref{eq:constraint}), and thus it will nonetheless be selected as the next token despite its low score.}.

\citet{li2022contrastive} set the score of the rest of the tokens in the vocabulary -- those not included in $\vhead$ -- to minus infinity. We argue that this design decision has the disadvantage of practically ignoring a large portion of the vocabulary, and thus losing information that can be useful. 

For instance, search-based decoding algorithms that rely on $S(x_t{\mid}x_{<t})$ will be limited to considering a small subset of the possible next tokens. Additionally, applications that require comparing the probabilities of a predefined and closed set of token options (See \citealp{liu2023pre}), will similarly lose valuable and pertinent information that was initially available in the LM$_{\textrm{orig}}$ probability distribution.

Thus, in \acd{} we retain the probabilities of the tokens not included in $\vhead$, keeping the distribution of the expert head:
\begin{equation}
S_\textsc{acd}(x_t{\mid}x_{<t})\label{eq:acd_full_a}\\ 
\!=\! \begin{cases} S(x_t{\mid}x_{<i})
& \text{if}~x_t\!\in\!\vhead \\
\pexp    	      & \text{otherwise} 
\end{cases}
\end{equation}

We further transform this score function into a probability distribution. The distribution of the expert head is split into two probability masses; one for the tokens in $\vhead$, and another for the tokens not included in it. We redistribute the former probability mass, weighted by the scores given to each token by Eq.~\ref{eq:cd}:
\begin{align}
& S_\textrm{redist}(x_t{\mid}x_{<t}) = \\
& \Softmax\Bigl(
    S(x_t{\mid}x_{<t})\Bigr) \cdot \sum_{x_t \in \vhead} p_\textsc{exp}(x_t{\mid}x_{<t})) \nonumber
\end{align}

Replacing $S(x_t{\mid}x_{<t})$ with $S_\textrm{redist}(x_t{\mid}x_{<t})$ in Eq.~\ref{eq:acd_full_a}, we obtain our auto-contrastive decoding probability distribution:
\begin{equation}
p_\textsc{acd}(x_t{\mid}x_{<t}) \label{eq:acd_full}
 \!=\! \begin{cases} S_\textrm{redist}(x_t{\mid}x_{<t})
& \text{if}~x_t\!\in\!\vhead \\
\pexp    	      & \text{otherwise} 
\end{cases}
\end{equation}

To summarize, auto-contrastive decoding, ACD, is a method to apply contrastive decoding over a single model. In \S \ref{ssec:heads} we explain how to create the amateur by adding and training a new head over an intermediate layer. In \S \ref{ssec:autocontrastive} we describe how to obtain a new probability distribution for the next token by contrasting the expert and the amateur.

\begin{figure*}
    \centering
    \begin{subfigure}[b]{0.45\textwidth}
        \includegraphics[width=\textwidth]{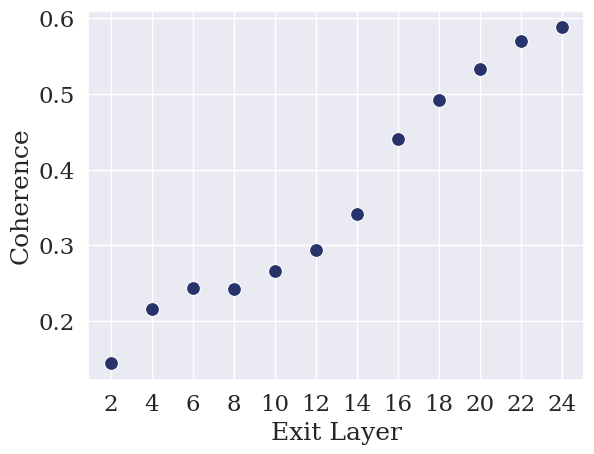}
        \caption{Coherence}
        \label{fig:coherence}
    \end{subfigure}
    \begin{subfigure}[b]{0.45\textwidth}
        \includegraphics[width=\textwidth]{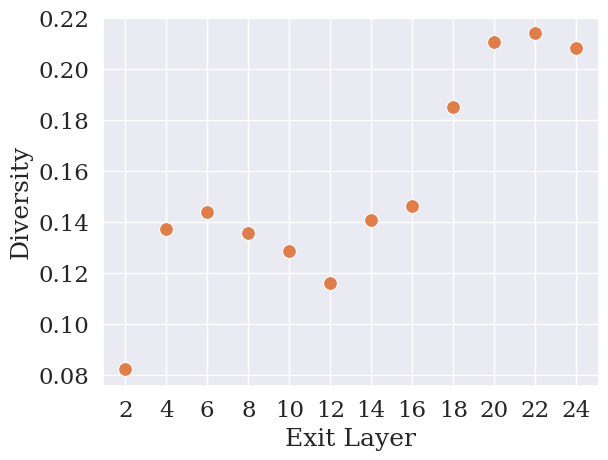}
        \caption{Diversity}
        \label{fig:diversity}
    \end{subfigure}
    \caption{\textbf{Open-ended generation for different exit layers.} The plots depict greedy decoding results of a pre-trained \med{} model, using different exit layers for generation. Each point represents an average over the \wikitext{} test examples of the coherence (a) and n-gram diversity (b).} \label{fig:coherence_diversity}
\end{figure*}
\section{Experimental Setup}
To test our approach, we conduct experiments on open-ended text generation, as well as on general language modeling benchmarks, comparing various performance metrics with and without applying auto-contrastive decoding. 

In order to analyze changes in performance across model layers, we add multiple new linear exit heads; thus, we also report and compare the baseline model behavior at different exit layers. 

\subsection{Models}
We use pre-trained auto-regressive language models from the GPT family -- GPT-2 \citep{radford2019language} and GPT-Neo -- as test models for exploring multi-exit performance and the effects of \acd. Specifically, we use the GPT-2 Medium ($355$M parameters, $24$ layers) and GPT-Neo-125M ($125$M parameters, $12$ layers) pre-trained model checkpoints\footnote{\small \url{https://huggingface.co/gpt2-medium}, \url{https://huggingface.co/EleutherAI/gpt-neo-125M}}.

As outlined in \S\ref{ssec:heads}, we create multi-exit variants of these models, that are identical to the original pre-trained checkpoints, other than the newly-added parameters for several new linear exit heads. To present a more comprehensive analysis, we add multiple heads, one connected to each of the even-numbered layers; thus, we add a total of $12$ and $6$ exit heads to \med{} and \neo{}, respectively. Each head uses the same configuration as the original language modeling head, with outputs for the $50257$ tokens in the vocabulary and an input size of $1024$ (\textit{GPT-2}) or $768$ (\textit{\neo{}}).

We train these heads on language modeling using self-supervision over the CC-$100$ \cite{conneau-etal-2020-unsupervised} corpus, following a standard pre-training approach (see Appendix~\ref{app:pretraining} for further details), keeping the original model parameters frozen. 
As described in \S\ref{ssec:heads}, when training the heads we do not precisely replicate the original pre-training regime; specifically, we use different pre-training data and train for a smaller number of training steps\footnote{for GPT-2, both the training corpus, and a comprehensive description of training details for the original pre-training, have not been publicly released; \neo{} was originally trained for $572$,$300$ steps over $300$ billion tokens.}. Nevertheless, we verify the quality of the training process by comparing the performance of a newly-trained final layer exit head to that of the original exit head of the pre-trained model (cf. App.~\ref{app:original_head}).

The pre-trained multi-exit base models are used as-is for open-ended text generation and for the benchmarks reported in \S\ref{ssec:benchmarks}. Model training and text generation were performed using the Hugging Face transformers library (v4.22.2) with the pytorch machine learning framework (v1.11.0).

\subsection{Tasks and Metrics}
\subsubsection{Open-ended generation} \label{ssec:open_ended_metrics}
Following \citet{li2022contrastive}, we evaluate  open-ended text generation in $3$ domains: books, Wikipedia, and news, using the BookCorpus \citep{zhu2015aligning}, \wikitext{} \citep{merity2017pointer}, and Wikinews\footnote{\small \url{http://www.wikinews.org}} text corpora, respectively. We test open-ended passage continuation by using the first $32$ words of a passage as a prompt, and using the multi-exit variant of the pre-trained model to decode up to $100$ tokens\footnote{\citet{li2022contrastive} decode $256$ tokens in continuation to the prompt, however they use stronger base models. With our models, generation deteriorates massively at those lengths.}.

Since \acd{} outputs a full probability distribution (see \S\ref{ssec:autocontrastive}), it can more naturally be combined with various existing decoding strategies. In this study we combine \acd{} with the following decoding methods: \textbf{Greedy search},
\textbf{Beam search} (\citealp{freitag2017beam}; \textit{beam=$5$}), 
\textbf{Top-k sampling} (\citealp{fan-etal-2018-hierarchical},  \textit{k=$50$}), 
and \textbf{Nucleus (top-p) sampling} (\citealp{holtzman2019curious}; \textit{p=$0.95$}). 

\begin{table}
\resizebox{\columnwidth}{!}{
\begin{tabular}{@{}lllllll@{}}
\toprule
\textbf{} & \multicolumn{2}{c}{\textbf{wikitext}}             & \multicolumn{2}{c}{\textbf{wikinews}}             & \multicolumn{2}{c}{\textbf{bookcorpus}}           \\ \midrule
\textbf{} & \multicolumn{1}{c}{\textit{div}} & \multicolumn{1}{c}{\textit{coh}} & \multicolumn{1}{c}{\textit{div}} & \multicolumn{1}{c}{\textit{coh}} & \multicolumn{1}{c}{\textit{div}} & \multicolumn{1}{c}{\textit{coh}} \\
\textbf{Greedy}      & 0.21 &          0.59 & 0.23 &          0.57 & 0.14 &          0.40 \\
\textbf{Greedy+\acd}  & \textbf{0.75} & \textbf{0.63}  & \textbf{0.74} & \textbf{0.61} & \textbf{0.62} & \textbf{0.50} \\ \\ \midrule
\textbf{Beam-5}      & 0.20 &          0.61 & 0.24 &          0.60 & 0.08 &          0.35 \\
\textbf{Beam-5+\acd}  & \textbf{0.57} & 0.62 & \textbf{0.58} & 0.61 & \textbf{0.37} & \textbf{0.48} \\ \\ \midrule
\textbf{Top-$k$}     & 0.96 &          0.57 & 0.96 &          0.55 & 0.97 &          0.42 \\
\textbf{Top-$k$+\acd} & 0.96 & \textbf{0.61} & 0.96 & \textbf{0.59} & 0.96 & \textbf{0.47} \\ \\ \midrule
\textbf{Top-$p$}     & 0.98 &          0.50 & 0.98 &          0.49 & 0.98 &          0.36 \\
\textbf{Top-$p$+\acd} & 0.98 & \textbf{0.55} & 0.98 & \textbf{0.54} & 0.98 & \textbf{0.41} \\ 
\end{tabular}
}
\caption{\textbf{The effect of \acd{} on open-ended generation.} This table lists the automatic quality metrics of n-gram diversity (\textit{div}) and topic coherence with the prompt (\textit{coh}) of a pretrained \med{} model, using different decoding strategies. For each strategy we compare results using the probability distribution of the exit head of the final ($24$th) model layer, to those obtained using an \acd{} probability distribution, contrasting the final layer next-token predictions with those of exit layer $12$.
}
\label{tab:open-ended-generation}
\end{table}

Generation quality is evaluated using automatic metrics focusing on different axes: aggregated \textbf{n-gram diversity} measures the repetitiveness within the generated continuations;
\textbf{semantic coherence} estimates topic drift by calculating similarity between the prompt and continuation. 
For further details on these metrics, refer to \citet{su2022contrastive}.

We also report results of human evaluation of the generation quality, comparing a sample of generation results across different settings, as explained below in \S\ref{ssec:open-gen}.

\subsubsection{Language modeling benchmarks}
We consider the pre-trained multi-exit model, which applies \acd{} at inference time and outputs complete next-token probability distributions (see \S\ref{ssec:autocontrastive}), to be a fully functional language model. This model contains the same parameters as \textit{LM$_{\textrm{orig}}$} (apart from the added linear exit heads), but differs in its characteristics. 

We therefore evaluate the \acd-enhanced model as a pre-trained language model, according to benchmarks that are commonly used (e.g., \citealp{black2022gpt, zhang2022opt}) to measure language modeling capabilities.

\textbf{LAMBADA} \citep{paperno2016lambada} is a popular benchmark that was proposed to encourage computational language models to keep track of information in the broader discourse, rather than paying attention to local context only. It has been shown that language models which exploit the context in a shallow manner perform poorly on this benchmark \citep{paperno2016lambada}. It is thus a relevant measure of more advanced language understanding abilities.

The typical measure used for reporting progress in language modeling is \textbf{Perplexity} \cite{jelinek1977perplexity}, the inverse of the (geometric) average probability assigned to each word in the test set by the model. Perplexity is commonly used as a measure of model quality, due in part to its simplicity and its relation to the maximum likelihood framework.
 
For running the benchmark tests, we use the Language Model Evaluation Harness library\footnote{\small \url{https://github.com/EleutherAI/lm-evaluation-harness}} (v0.3.0).

\begin{table*}
\resizebox{\textwidth}{!}{
\begin{tabular}{p{0.1\linewidth} | p{0.22\linewidth} | p{0.33\linewidth} | p{0.35\linewidth}}
\hline
\textbf{\small{Mitigated failure}} & \textbf{\small{Prompt}}                                                                   & \textbf{\small{Greedy \acd}}                                                                                                                                                                      & \textbf{\small{Greedy}}                                                                                                                                                                                     \\ \hline
\textit{\small{Short loop}}        & \small{The use of iron instead of wood as the primary material of}& \small{furniture could have created problems, says study leader Prof. Iain Kelly from the School of Materials Science and Engineering at the University}                              & \small{the building blocks of the building blocks of the building blocks of the building blocks of the building blocks of the building blocks of the} \\ \hline

\textit{\small{Longer loop}}       &\small{Du Fu's political comments are based on emotion rather than calculation:}                     &\small{if his party loses power, he fears, China will face an even more uncertain future than it faces now. He fears a}                       
& \small{he is a man who has been in the trenches for years, and he is a man who has been in the trenches for}
\\ \hline

\textit{\small{Prompt repeated}} & \small{The first ironclads to have all-steel armor were the Italian Caio Duilio}                & \small{in 1230 and the Saxon Magnus in 1252, both of whom wore steel shields. Iron armor became so common that}
& \small{and the German Wilhelm von Habsburg. The first ironclads to have all-steel armor were the Italian Caio}                                                    
\\ \hline
\end{tabular}
}
\caption{\textbf{Examples of common failures diminished when applying \acd.}}
\label{tab:failures}
\end{table*}

\section{Results and Analysis} \label{sec:results}

\subsection{Open-ended generation} \label{ssec:open-gen}
Results for open-ended text generation for the \textit{\med{}} model are shown in Table~\ref{tab:open-ended-generation}. For the \textit{greedy} and \textit{beam-search} strategies, which exhibit low diversity of generated texts, we see a significant improvement in diversity when combining them with \acd.
At the same time, semantic coherence scores with \acd{} are higher in almost all settings tested. 
Similar effects of \acd{} can be observed for the smaller \neo{} model (App. Table~\ref{tab:open-ended-generation_neo}).

The gains in text diversity highlight one major effect of \acd, which is that of reducing repetitiveness in generation. This is true both to short loops, such as two tokens being generated again and again, as well as longer ones. Also, in some cases texts generated by the top layer simply repeat/variate the prompt. See Table~\ref{tab:failures} for examples of the above failures and their mitigation.
% Note that in the latter case, of the prompt being repeated precisely or with slight modifications, in terms of \textit{coherence} the generated output would not be punished by the automatic metric, but rather receive a high coherence score.

\begin{table}
\begin{tabular}{@{}lcc@{}}
\toprule
\textbf{} & \multicolumn{1}{c}{\textbf{Diversity}} & \multicolumn{1}{c}{\textbf{Coherence}}\\
\midrule
GPT2-Medium     & 0.22 & 0.63 \\
GPT2-XL      & 0.31 & 0.63 \\
\midrule
\textit{GPT2-Medium + \acd}    & \textbf{0.75} & 0.63 \\
\bottomrule
\end{tabular}

\caption{\textbf{Scale effects of \acd{} on open generation.} Depicted are topic coherence and n-gram diversity of generation outputs over WikiText-103, for $3$ settings: a large model (\xl{}, $1.5$B parameters), a medium-sized model (\med{}, $355$M parameters, using its original exit head), and the same medium-sized model applying \acd{} at inference time, contrasting the next-token predictions of the final ($24$th) layer and layer $12$.}
\label{tab:xl_open-ended-generation}
\end{table}

Given the dramatic performance boost given by \acd, as seen in Tables~\ref{tab:open-ended-generation} and \ref{tab:open-ended-generation_neo}, we further ask how \acd-enhanced generation outputs would compare to those of a \textit{larger model} with more advanced capabilities. To this end, we perform open-ended generation using the \textit{GPT2-XL} model (1.5B parameters). As can be seen in Table~\ref{tab:xl_open-ended-generation}, \med{} (355M parameters) that is enhanced by \acd{} significantly outperforms its larger scale counterpart.

To verify that these results are robust and not an artifact of the automatic measures used, we conduct human evaluation of a sample of generation outputs from the results in Table~\ref{tab:xl_open-ended-generation}, presenting the prompt and pairs of generated texts to human annotators and asking them to compare the quality of outputs. Results indicate that outputs from \xl{} were twice as likely to be judged as better compared to the baseline \med{}; but strikingly, \med{} outputs obtained using \acd{} were overall judged as slightly better than those of the much larger \xl{}. For details on the human evaluation task, refer to App.~\ref{app:human_eval}.

In Fig.~\ref{fig:coherence} we portray the behavior of the automatic coherence measure when relying on the outputs of different \textit{\med{}} model exits. It appears that the generation coherence, i.e., the semantic relatedness between the prompt and generated continuation, rises consistently as progressing from lower to higher layers. Presumably, this reflects a gradual decrease in topic drift behaviors and an increased ability to generate longer sequences that remain semantically coherent.

Fig.~\ref{fig:diversity} depicts the diversity of open-ended generation across layers. Interestingly, this measure exhibits more complex patterns, rising and falling as we progress from lower to higher layers. As is common with automatic quality metrics for text generation, we see this as an indication that n-gram repetition provides only a partial window into the generation quality, particularly where the diversity is overall quite low. Moreover, the nature of outputs may undergo phase shifts as they improve. For instance, generated sequences may shift from being diverse but unrelated to the inputs in lower layers, to texts that are semantically related to the prompt but highly repetitive, and so on.

\begin{figure*}
    \centering
    \includegraphics[width=0.45\textwidth, page=1]{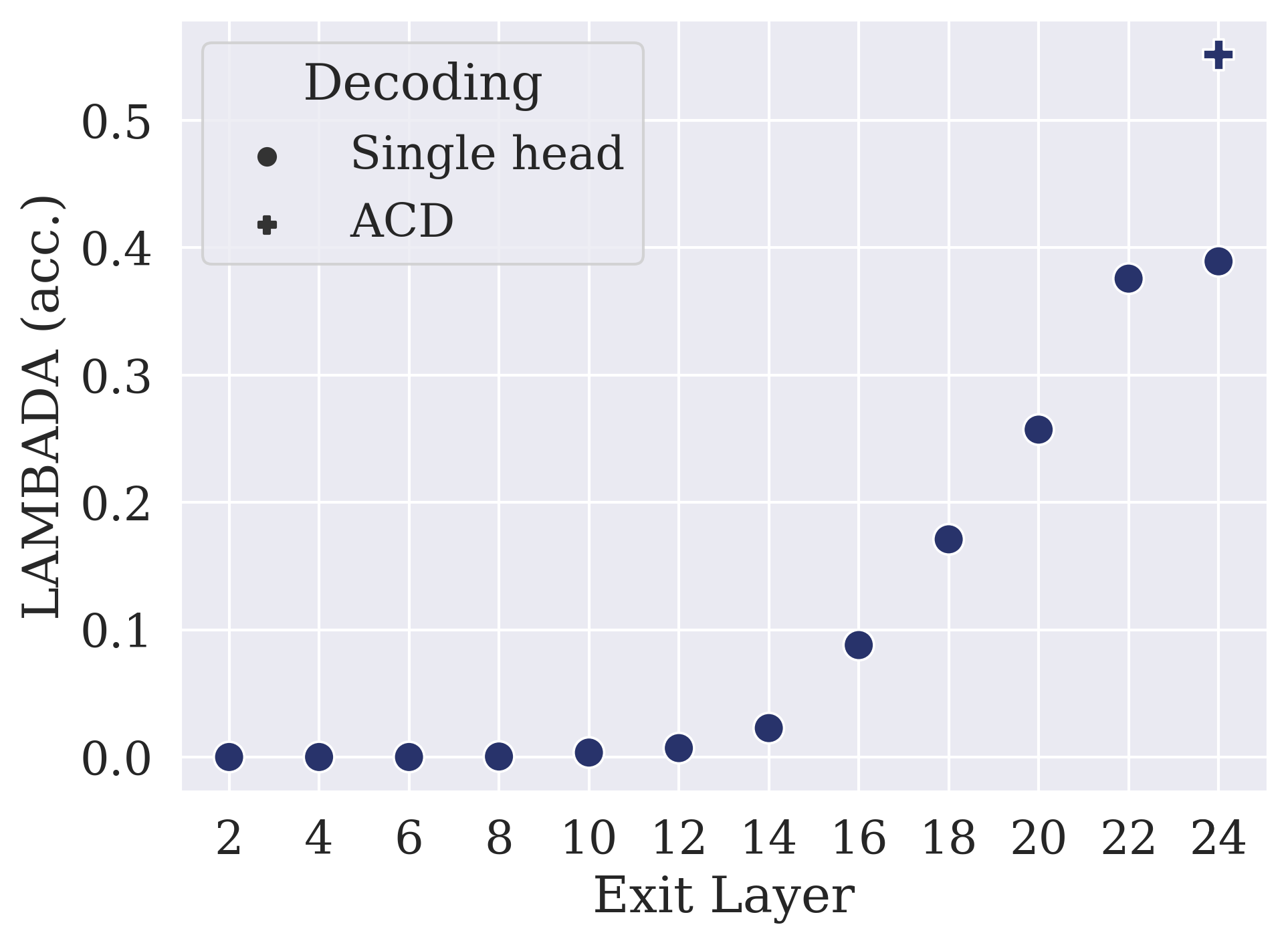} 
    \includegraphics[width=0.45\textwidth, page=2]{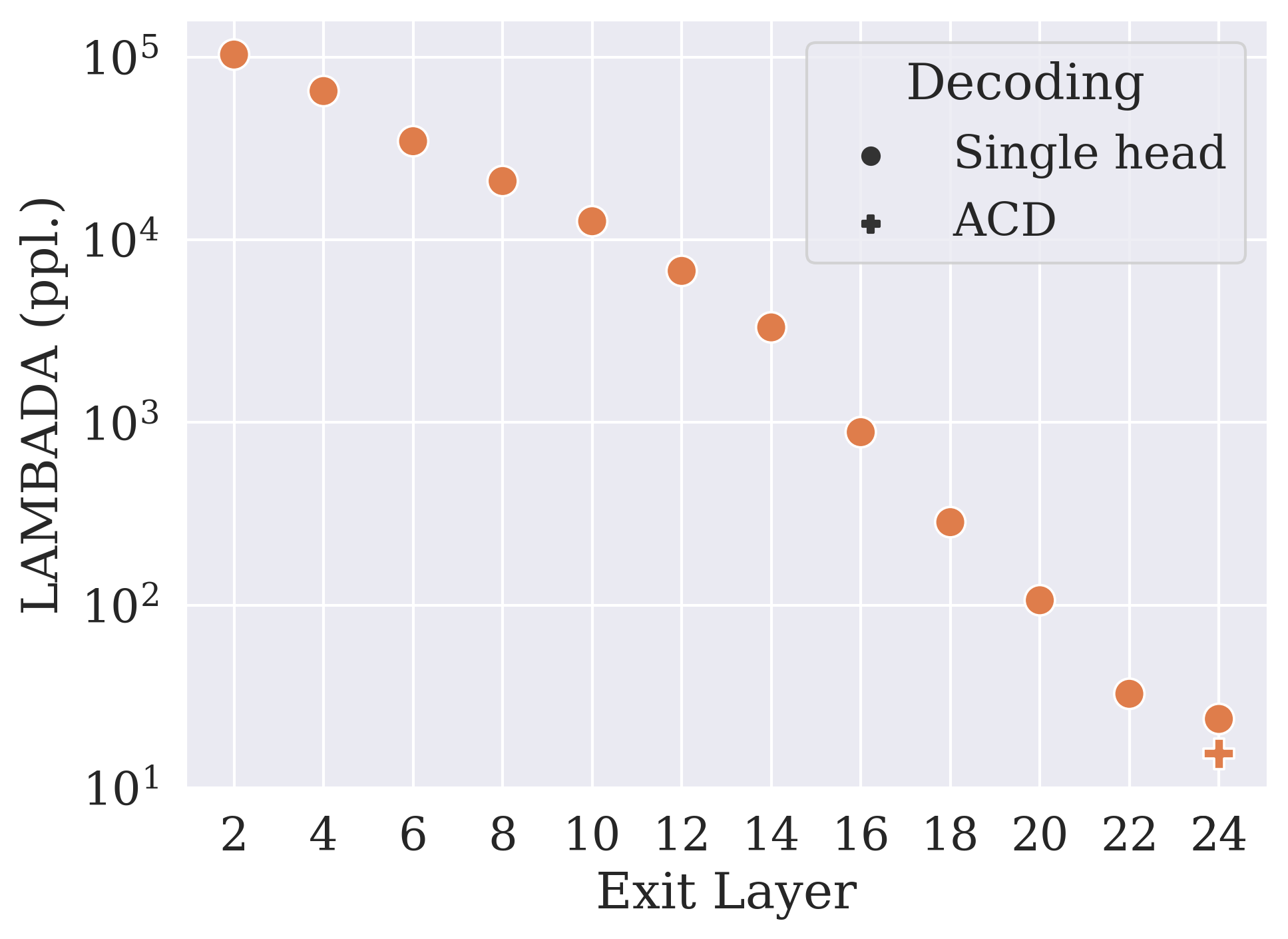}
     \caption{\textbf{GPT2 performance on the LAMBADA benchmark}. Plots depict accuracy (left, higher is better) and perplexity (right, lower is better; presented in log scale) on LAMBADA 
     across layers. Individual \textit{\med{}} exits are denoted by •; results for the \acd{} probability distribution, contrasting layers $24$ and $12$, are denoted by +. \label{fig:lambada}
    }
\end{figure*}

\subsection{Language modeling benchmarks} \label{ssec:benchmarks}
Results for the LAMBADA benchmark task, for individual exit layers of \med{} and for \acd{} generation, are shown in Figure~\ref{fig:lambada}. The accuracy and the perplexity metrics of this benchmark dataset both improve as progressing along the model layers. In both cases, performance is further improved by applying \acd, with substantial gains in accuracy. Similar gains are obtained for the \neo{} model (App. Figure~\ref{fig:lambada_neo}).

This is a non-trivial finding, in that it provides an indication that by using \acd{} we enable the model to more accurately take into account the broader context and long-range dependencies in the text.

As in \S\ref{ssec:open-gen}, one may further ask how these gains compare to the performance reached by a larger pre-trained model. Indeed, as shown in Table~\ref{tab:xl_lambada}, \med{} enhanced by \acd{} is on par with the larger \textit{GPT2-XL} model (1.5B parameters) on the LAMBADA benchmark, achieving improved accuracy but also somewhat inferior perplexity.

\begin{figure}[th]
\begin{center}
\includegraphics[width=\columnwidth]{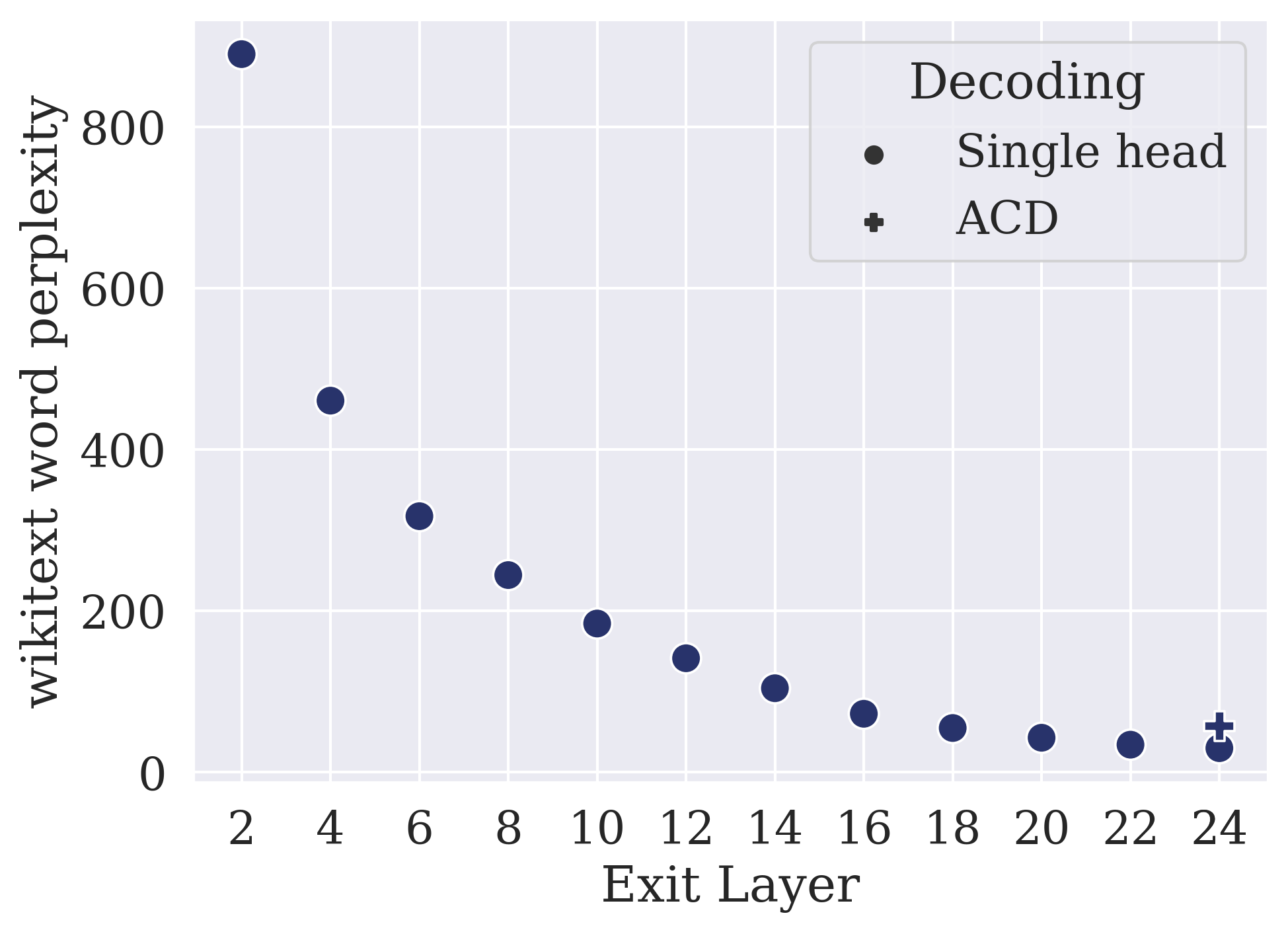}
\setlength{\belowcaptionskip}{-20pt}

\caption{\textbf{GPT2 performance on the WikiText-2 benchmark}. The plot depicts the word-level perplexity (lower is better) over the WikiText-2 corpus. Individual \textit{\med{}} exit layers are denoted by •, while results for the \acd{} probability distribution, contrasting layers $24$ and $12$, are denoted by +.}
\label{fig:wikitext_ppl}
\end{center}

\end{figure}

Figure~\ref{fig:wikitext_ppl} depicts the word-level perplexity over the general WikiText-$2$ dataset.
% \footnote{\url{https://www.salesforce.com/products/einstein/ai-research/the-wikitext-dependency-language-modeling-dataset/}}. 
As can be seen, perplexity behaves as expected across model layers. For this general corpus, \acd{} does not improve the overall perplexity beyond that of the final exit layer.

Thus, we see that \acd{} provides a substantial benefit for the challenging LAMBADA data, that specifically measures a model's advanced ability to look at broader context windows, but not for the overall perplexity over a general text corpus. While this is an initial finding that deserves further exploration, one interpretation is that \acd{} specifically strengthens "higher-layer behaviors", such as those measured by the challenging LAMBADA task, but also induces other types of biases into the model's output probability distributions.

\section{Discussion} \label{sec:discussion}

In this work we develop an approach that contrasts different model layers, improving the output probabilities of a generative model. Applying it to existing pre-trained language models, we demonstrate that intermediate low-performing model layers can in some cases inform the predictions of the high-performance final layer. This setting is of particular interest due to its practicality and flexibility, as it can be applicable to models of different sizes and is utilized during inference via a single forward pass. 

\begin{table}
\resizebox{\columnwidth}{!}{
\begin{tabular}{@{}lcc@{}}
\toprule
\multicolumn{1}{r}{\textit{LAMBADA}} & \multicolumn{1}{c}{\textbf{Acc. $\uparrow$}} & \multicolumn{1}{c}{\textbf{Ppl. $\downarrow$}}\\
\midrule
GPT2-Medium     & 0.43 & 18.3 \\
GPT2-XL      & 0.51 & \textbf{10.6} \\
\midrule
\textit{GPT2-Medium + \acd}    & \textbf{0.55} & 15.4 \\
\bottomrule
\end{tabular}
}
\caption{\textbf{Scale effects of \acd{} on LAMBADA.} Depicted are the accuracy and perplexity scores of the LAMBADA benchmark, for $3$ settings: a large model (\xl{}, $1.5$B parameters), a medium-sized model (\med{}, $355$M parameters, using its original exit head), and the same medium-sized model applying \acd{} at inference time, contrasting the next-token predictions of the final ($24$th) layer and layer $12$.}
\label{tab:xl_lambada}
\end{table}

But more broadly, our findings bring forth an enticing notion, that one would be able to make more out of an existing model simply by considering the predictions of intermediate layers (which are typically ignored). This idea is somewhat counter-intuitive, as language models are in a sense optimized -- and often in a long pretraining process over massive corpora -- for the quality of their final layer representations. At the same time, thematically this notion is in line with works that describe the computations in transformer models as a linear-like progression, where each layer refines the representations of the previous ones, and where even the representations of specific tokens can shift in a consistent direction along with the progression across layers \citep{geva-etal-2021-transformer, geva2022transformer}. Loosely speaking, if the changes from one layer to the next can sometimes track a vector of improvement with a discernible direction, then in theory one could try and "extend" this vector; and doing so may help estimate of what a \textit{larger} model, one with additional layers, would have said about a particular instance. We see these as interesting avenues both for theoretical study, and for empirical explorations as to whether surprising findings such as those presented here can be applied to real-world use-cases.

Here we present an initial, and relatively simple, algorithm for performing the \acd{} contrast between layers. As in \citet{li2022contrastive}, our formulation still relies on a somewhat arbitrary hyperparameter $\alpha$; also, contrast is always done with respect to a single particular exit layer, and choosing the most appropriate layer for contrast may not be trivial. 
Here, for simplicity and robustness, we did not attempt to optimize these two important hyper-parameters, and used a single configuration throughout our experiments. However, we see much room  for future work on improving these details, and finding ways to intelligently choose which layers to contrast and how to combine between them.

An interesting avenue for future work concerns the effect of \acd{} when applied not just to a pre-trained model, but to one fine-tuned for a particular downstream task. Specifically, it may be that specific types of generation tasks may derive more benefit from \acd, depending on their reliance on more "high-level" model capabilities, and also on the importance of diversity in generated outputs.

The present work focuses specifically on generative models, and on improving the quality of text generation outputs and next-token predictions. However, the basic approach of looking at the outputs of intermediate layers and using them to inform model predictions is a general one, and is thus also worth exploring in other contexts, such as classification tasks.

To sum, our findings indicate that our proposed approach, \acd, can be of great practical value, in that it significantly boosts the performance of a generative language model with a minimal computational cost. This approach suggests new avenues on how to best extract knowledge from a language model and more efficiently utilize its parameters.

\section*{Limitations}
One of the primary limitations of this work is that this is essentially an empirical study. Although we provide extensive experiments to show that the proposed approach demonstrates significantly better results in different settings, currently we do not provide any theoretical guarantees for this approach. Second, many of our experiments would not be easily reproduced in languages other than English, that lack sufficient linguistic resources. During this study we used the \textit{GPT-2} and \textit{GPT-Neo} language models, which have been trained on large amounts of English text. Finally, anecdotally we observed that this approach can also increase hallucination behaviors, which are a common issue with many text generation models. During application, one would have to take necessary measures to monitor the hallucinations produced by the model.

\section*{Acknowledgements}
We thank our many colleagues for their valuable input on this research effort, and owe particular thanks to Liat Ein-Dor and Leshem Choshen for their advice and assistance.

\bibliography{custom}
\bibliographystyle{acl_natbib}

\appendix
\section{Pre-training details} \label{app:pretraining}
For training the additional linear heads in our multi-exit versions of \med{} and \neo{}, we apply a training regime to the pre-trained models, while freezing the parameters of the original pre-trained model checkpoints (see \S\ref{ssec:heads}).

For runtime considerations, we train all the added linear heads ($12$ and $6$ heads in total for \med{} and \neo{}, respectively) within a single training run, where a cross-entropy loss is calculated for the outputs of each individual linear head with respect to the labels, and the total training loss is calculated as the sum of these losses. Note that since each head is only connected to its exit layer $m$, and the shared pre-trained model parameters are kept frozen, this setup is roughly equivalent to training each of the linear heads separately.

Training was conducted with self-supervision over the English portion of the CC-$100$ \citep{conneau-etal-2020-unsupervised} corpus\footnote{\url{https://huggingface.co/datasets/cc100}}. We used $20$M instances out of the full dataset. Each text was tokenized, and the different tokenized instances were then joined together into chunks with a maximum sequence length of $512$. Thus, no padding was applied to the examples. Following the tokenization and chunking, the training data consisted of $\sim{1.3}$M training examples ($\sim{650}$M tokens).
Training was performed using a causal language modeling objective, where the cross-entropy loss is calculated between the autoregressively generated outputs of the language modeling head and the input tokens (of length $512$), which serve as the label.

The linear heads of each model were trained for $3$ epochs over the chunked texts, using the AdamW optimizer, a learning rate of $2\times10^{-4}$ with a linear decay scheduler, and a train batch size of $64$.

\section{\neo{} results}
\begin{figure*}
    \centering
    \includegraphics[width=0.48\textwidth, page=1]{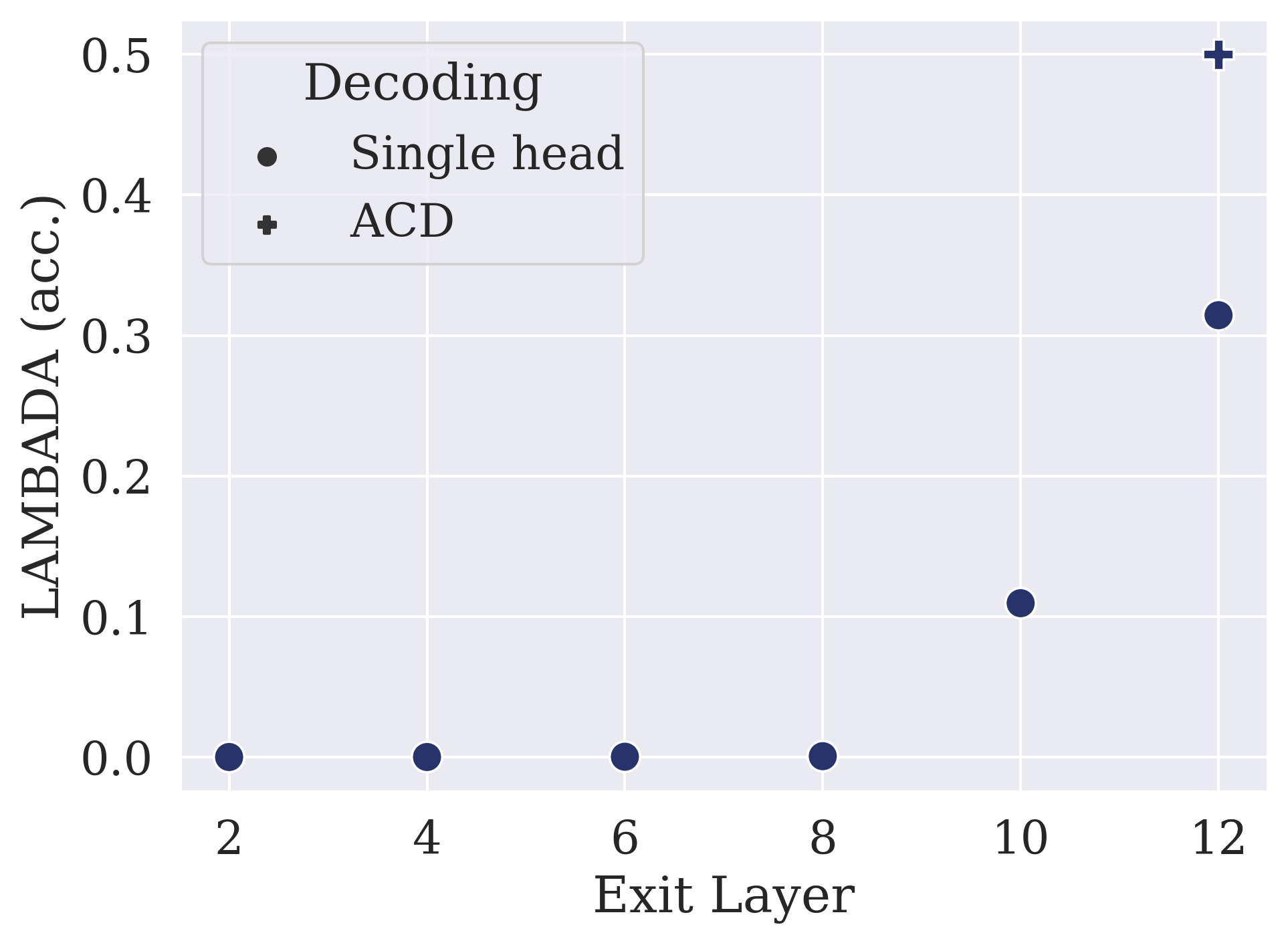} 
    \includegraphics[width=0.48\textwidth, page=2]{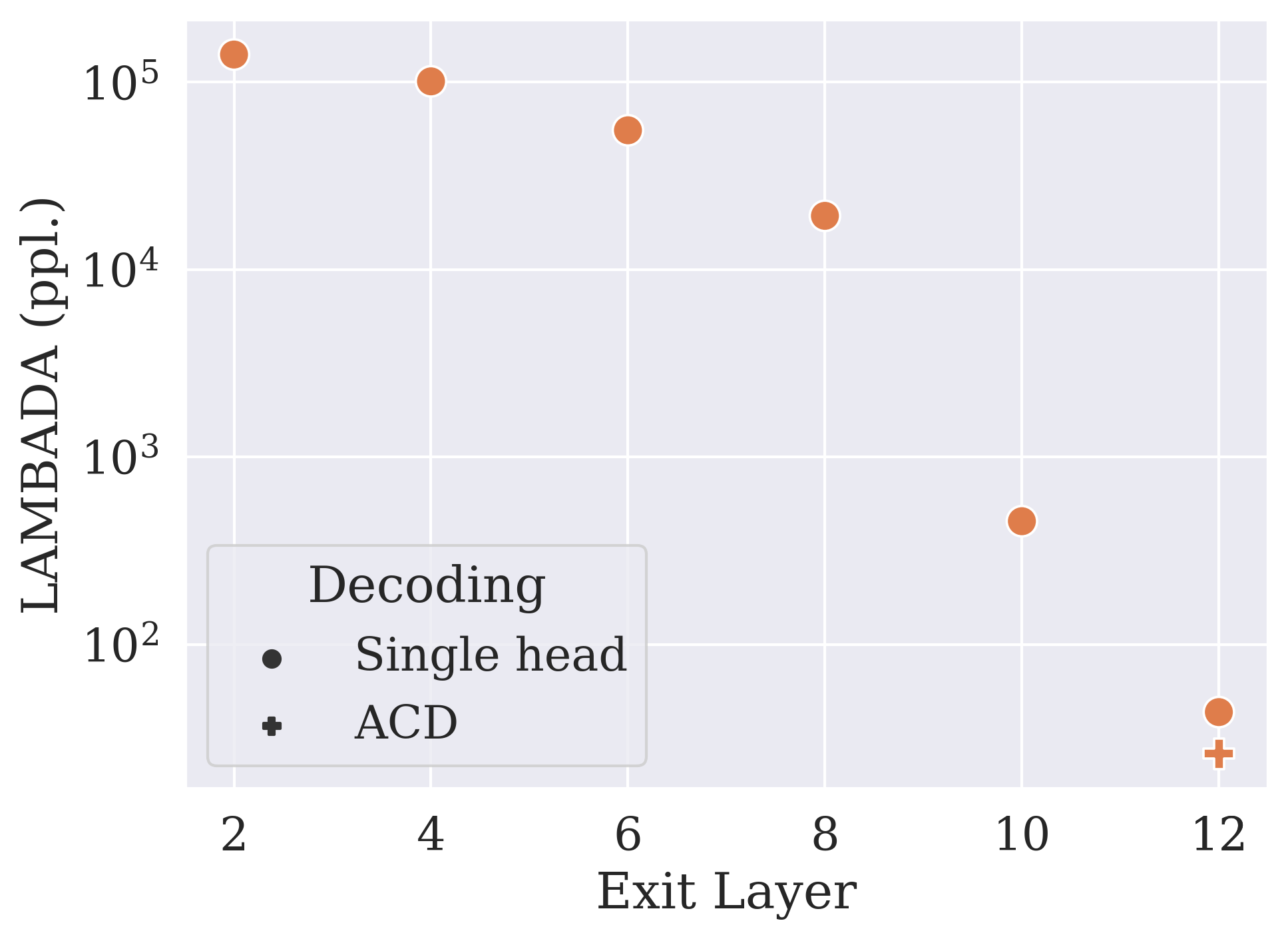}
     \caption{\textbf{GPT-Neo performance on the LAMBADA benchmark}. Plots depict accuracy (left, higher is better) and perplexity (right, lower is better; presented in log scale) on the LAMBADA language modeling task across different layers. Individual \textit{GPT-Neo-125M} exit layers are denoted by •, while results for the \acd{} probability distribution, contrasting layers $12$ and $8$, are denoted by +. \label{fig:lambada_neo}
    }
\end{figure*}

The open-generation results for the \neo{} model are shown in Table~\ref{tab:open-ended-generation_neo}. The results for this model over the LAMBADA benchmark are depicted in Fig.~\ref{fig:lambada_neo}.

\begin{table}
\resizebox{\columnwidth}{!}{
\begin{tabular}{@{}lllllll@{}}
\toprule
\textbf{} & \multicolumn{2}{c}{\textbf{wikitext}}             & \multicolumn{2}{c}{\textbf{wikinews}}             & \multicolumn{2}{c}{\textbf{bookcorpus}}           \\ \midrule
\textbf{} & \multicolumn{1}{c}{\textit{div}} & \multicolumn{1}{c}{\textit{coh}} & \multicolumn{1}{c}{\textit{div}} & \multicolumn{1}{c}{\textit{coh}} & \multicolumn{1}{c}{\textit{div}} & \multicolumn{1}{c}{\textit{coh}} \\
\textbf{Greedy}      & 0.09 &          0.57 & 0.08 &          0.54 & 0.06 &          0.35 \\
\textbf{Greedy+\acd}  & \textbf{0.32} & \textbf{0.62}  & \textbf{0.32} & \textbf{0.61} & \textbf{0.20} & \textbf{0.49} \\ \\ \midrule
\textbf{Beam-5}      & 0.08 &          0.59 & 0.08 &          0.56 & 0.05 &          0.33 \\
\textbf{Beam-5+\acd}  & \textbf{0.15} & 0.60 & \textbf{0.15} & \textbf{0.60} & \textbf{0.10} & \textbf{0.48} \\ \\ \midrule
\textbf{Top-$k$}     & \textbf{0.95} &          0.56 & \textbf{0.95} &          0.54 & \textbf{0.95} &          0.40 \\
\textbf{Top-$k$+\acd} & 0.91 &          \textbf{0.62} & 0.92 &          \textbf{0.60} & 0.92 &          \textbf{0.48} \\ \\ \midrule
\textbf{Top-$p$}     & 0.98 &          0.48 & 0.98 &          0.47 & 0.98 &          0.35 \\
\textbf{Top-$p$+\acd} & 0.97 & \textbf{0.56} & 0.97 & \textbf{0.54} & 0.97 & \textbf{0.41} \\ 
\end{tabular}
}
\caption{\textbf{The effect of \acd{} on open-ended generation.} This table lists the automatic generation quality metrics of n-gram diversity (\textit{div}) and topic coherence with the prompt (\textit{coh}) of a pretrained \neo{} model, using different decoding strategies. For each strategy we compare results using the probability distribution of the exit head of the final ($12$th) model layer, to those obtained using an \acd{} probability distribution, contrasting the final layer next-token predictions with those of exit layer $8$.
}
\label{tab:open-ended-generation_neo}
\end{table}

\section{Comparison to the original LM heads}
\label{app:original_head}
As noted in \S\ref{ssec:heads}, in order to reduce training disparities between the expert and the amateur we train a new expert head, rather than using the model's original exit head as the expert. Here, we compare the performance of the newly-trained expert heads to that of the original language modeling heads. In addition, we report the effects of \acd{} when using the original expert head for the \acd{} contrast procedure.

As can be seen in Table~\ref{tab:original_head}, our newly-trained expert heads are slightly inferior to the original language modeling heads, presumably due to the more limited pre-training regime of the new heads. Nevertheless, \acd{} that relies on the newly-trained expert head clearly outperforms the original language modeling head in open-generation and LAMBADA metrics (as also shown in Tables~\ref{tab:xl_open-ended-generation} and \ref{tab:xl_lambada}).

The results of \acd{} when contrasting between the \textit{original} LM head and our newly-trained amateur head are overall rather similar. Thus, despite the more noisy or unpredictable nature of the disparities between the exit heads in this case (given that they were trained in a different pre-training regime over different training examples), it appears the effects of applying \acd{} are relatively robust to such a scenario.

\begin{table*}
\begin{tabular}{@{}lcccc@{}}
\toprule
 & \multicolumn{1}{c}{Diversity $\uparrow$} & \multicolumn{1}{c}{Coherence $\uparrow$} & \multicolumn{1}{c}{LAMBADA acc. $\uparrow$} & \multicolumn{1}{c}{Perplexity $\downarrow$}\\
\midrule
\midrule
\med{} \textit{L-$24_{\textrm{orig}}$}  & 0.22 & 0.63 & 0.43 &  26.8 \\
\med{} \textit{L-$24_{\textrm{new}}$} & 0.21 & 0.59 & 0.39 & 30.0 \\
\midrule
\med{} \textit{L-$24_{\textrm{orig}}$} + \acd{} & 0.62 & 0.64 & 0.56 & 71.3 \\
\med{} \textit{L-$24_{\textrm{new}}$} + \acd{} & 0.75 & 0.63 & 0.55 & 57.2 \\
\midrule
\midrule
\neo{} \textit{L-$12_{\textrm{orig}}$}  & 0.12 & 0.60 & 0.37 &  32.3 \\
\neo{} \textit{L-$12_{\textrm{new}}$} & 0.09 & 0.57 & 0.31 & 38.5 \\
\midrule
\neo{} \textit{L-$12_{\textrm{orig}}$} + \acd{} & 0.30 & 0.62 & 0.53 & 68.1 \\
\neo{} \textit{L-$12_{\textrm{new}}$} + \acd{} & 0.32 & 0.62 & 0.50 & 71.8 \\
\bottomrule
\end{tabular}

\caption{\textbf{Comparison to the original LM exit heads.} Depicted are open-generation metrics (using greedy decoding over WikiText-103), LAMBADA benchmark accuracy, and WikiText-2 perplexity of the \med{} and \neo{} models. For each model, $4$ settings are shown: using its original exit head (\textit{L-$_{\textrm{orig}}$}), using our newly-trained final layer exit head  (\textit{L-$_{\textrm{new}}$}), and the results of applying \acd{} at inference time, contrasting the next-token predictions of a newly-trained intermediate layer exit head with those of either the original (\textit{L-$_{\textrm{orig}}$ + \acd{}}) or newly-trained (\textit{L-$_{\textrm{new}}$ + \acd{}}) final layer exit.}
\label{tab:original_head}
\end{table*}

\section{Human evaluation} \label{app:human_eval}

\begin{figure*}[t]
\includegraphics[scale=0.25]{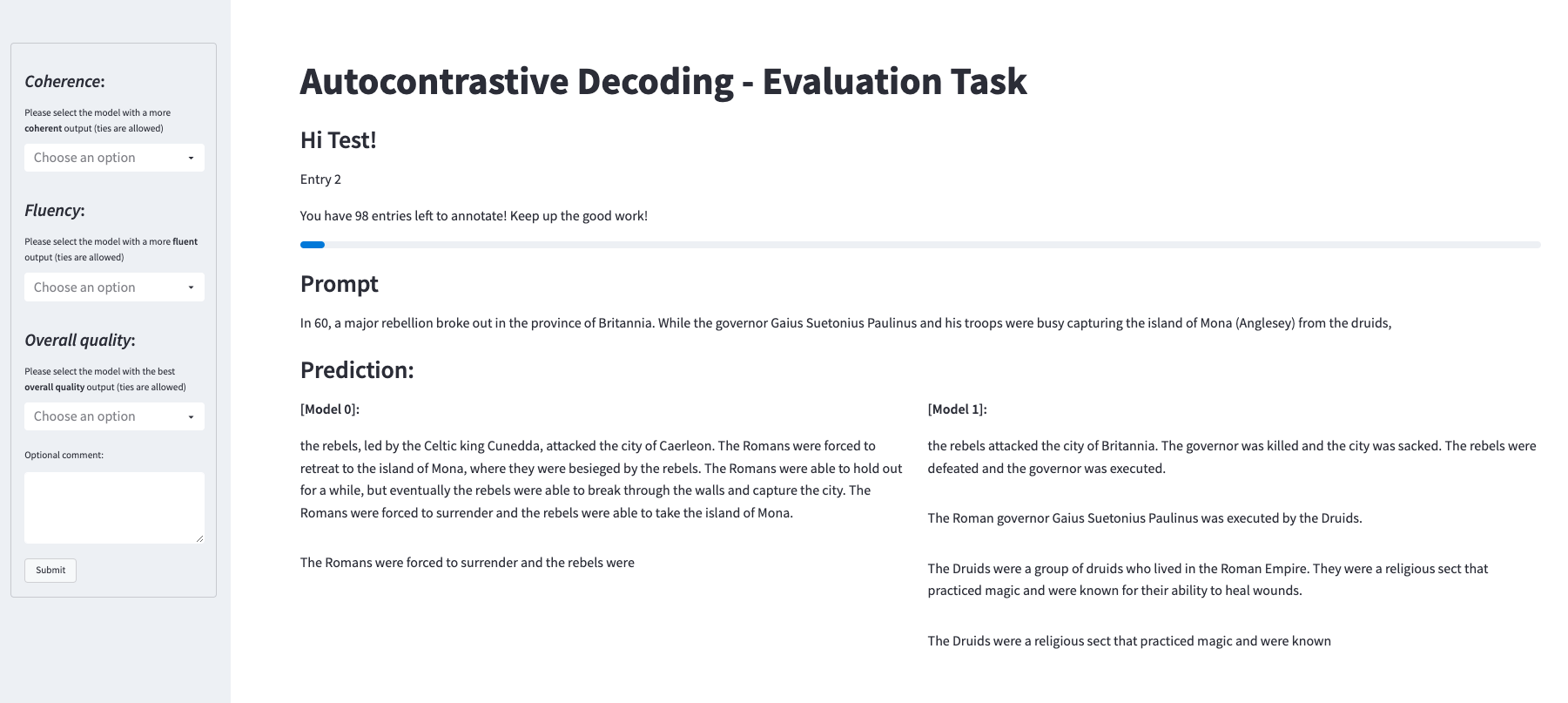}
  \caption{\textbf{Human Evaluation Task UI}}
  \label{fig:human_screenshot}
\end{figure*}

\begin{figure}[t]
\includegraphics[scale=0.4]{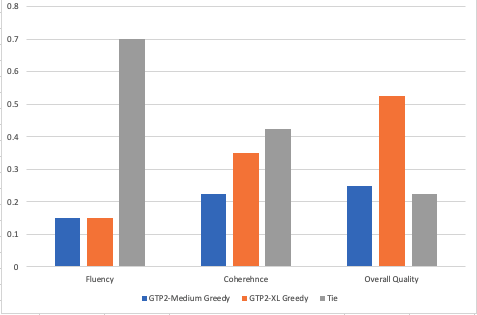}
  \caption{\textbf{Human Evaluation results comparing greedy decoding generation outputs of \xl{} and \med{}}. This figure shows the distribution of majority labels for each of the $3$ task questions. }
  \label{fig:human_large_med}
\end{figure}

\begin{figure}[t]
\includegraphics[scale=0.4]{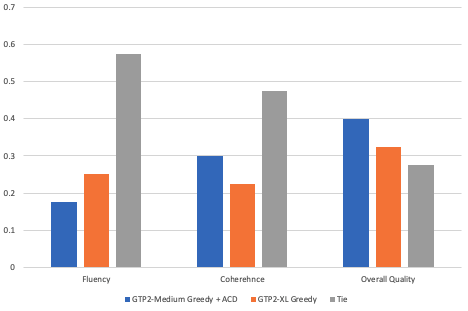}
  \caption{\textbf{Human Evaluation results comparing greedy decoding generation outputs of \xl{} and \med{} + \acd{}}. This figure shows the distribution of majority labels for each of the $3$ task questions.}
  \label{fig:human_large_acd}
\end{figure}

We conducted two evaluations for open-ended generation quality of the models:
\begin{itemize}
    \item{Comparing greedy decoding outputs of \textbf{\xl{}} and \textbf{\med{}}}
    \item {Comparing greedy decoding outputs of \textbf{\xl{}} to \textbf{\med{} with \acd{}}}
\end{itemize}

As input for inference, we randomly sampled $40$ texts from the \wikitext{} dataset. Following the setting described in \S\ref{ssec:open_ended_metrics}, we used the first $32$ words of those texts as prompts and for each evaluated model extracted up to $100$ tokens of the decoded text. The same prompts were used for the two sets of evaluations, and thus also identical generation outputs of the \xl{} Greedy setting.

$3$ NLP experts labeled the $80$ resulting instances, consisting of a prompt and inferences from two models. For each instance, they were asked to select the better model on $3$ different aspects, in separate questions: \textit{fluency}, \textit{coherence} and \textit{overall quality} (Figure~\ref{fig:human_screenshot}). For each question they could select either `model A', `model B' or a tie. The inferences were shuffled such that 'model A' for each displayed instance was randomly selected from either the \xl{} Greedy model or its counterpart. The  sets of evaluations (i.e., \xl{} vs. \med{} and \xl{} vs. \med{} + \acd{}) were also shuffled, such that annotators did not know which pair of models they are annotating.  

The final label for each instance is obtained by the majority choice of the annotators. A \textit{tie} majority label is achieved either when the majority of annotations is \textit{tie} or when no majority is obtained (which in this setting can only occur when annotations are equally distributed - one for each model and one \textit{tie}). 

Label distributions are shown in Figures \ref{fig:human_large_med}, \ref{fig:human_large_acd}. Inter-annotator agreement for those tasks, obtained by averaging Cohen's Kappa for all annotator pairs, in each task, for each question is as follows - $0.15$ for the fluency question, $0.34$ for the coherence question and $0.42$ for the overall quality question. An image of the task is shown in Figure \ref{fig:human_screenshot}.

\end{document}